\documentclass[a4paper,12pt]{article}

\usepackage{arxiv}

\usepackage[utf8]{inputenc} % allow utf-8 input
\usepackage[T1]{fontenc}    % use 8-bit T1 fonts
\usepackage{hyperref}       % hyperlinks
\usepackage{url}            % simple URL typesetting
\usepackage{booktabs}       % professional-quality tables
\usepackage{amsfonts}       % blackboard math symbols
\usepackage{nicefrac}       % compact symbols for 1/2, etc.
\usepackage{microtype}      % microtypography
\usepackage{lipsum}

\usepackage{multicol} %несколько колонок
\usepackage[pdftex]{graphicx}

\graphicspath{{noiseimages/}}
\usepackage{enumitem}
\usepackage{amssymb}

\usepackage{blindtext}
\usepackage{scrextend}
\addtokomafont{labelinglabel}{\sffamily}

\usepackage{amsmath}
\usepackage{enumerate}
\setlist{nolistsep, itemsep=0.1cm,parsep=1pt,leftmargin=0cm}
\usepackage{multirow}

\title{Neural network-based object classification\\ by known and unknown features\\ (based on text queries)}

\author{
  A.~Artemov,\ I.~Bolokhov,\ D.~Kem,\ I.~Khasenevich
   \\
  Cognitive Systems Company, BRAIN2 Project\\
  \texttt{science@cogsys.company} \\
}

\begin{document}
\maketitle

\begin{abstract}
The article presents a method that improves the quality of classification of objects described by a combination of known and unknown features. The method is based on modernized Informational Neurobayesian Approach with consideration of unknown features. The proposed method was developed and trained on 1500 text queries of Promobot users in Russian to classify them into 20 categories (classes). As a result, the use of the method allowed to completely solve the problem of misclassification for queries with combining known and unknown features of the model. 
The theoretical substantiation of the method is presented by the formulated and proved theorem On the Model with Limited Knowledge.  It states, that in conditions of limited data, an equal number of equally unknown features of an object cannot have different significance for the classification problem.
\end{abstract}
\begin{multicols}{2}
\textbf{Keywords:} Informational Neurobayesian Approach, Neural Networks, Unknown Features, Machine Learning, NLP

\section{Introduction}
Modern machine learning methods allow to classify various objects with high accuracy on the basis of specific features. However, for most of these methods, including neural networks, only known features are used to classify the object.

\begin{center}
\Large Known features $\rightarrow$ Object class
 
\normalsize  \textit{%Figure 1. 
Illustration of the connection between object’s features and class}
\end{center}

As a result, these methods classify objects with equal result their known features and with the addition of unknown features.  We consider this approach to be inadequate. This inadequacy can be demonstrated by the following example. Suppose that there is a phrase in a foreign language that describes an object. Most of the words (including their possible properties) of the phrase are unknown, except for the fragment – "the object has four wheels and a steering wheel." Suppose the task is to answer whether this phrase describes a car, a race car, or an excavator. Unknown words may bring us closer or farther from these categories. For example, "super-fast"- identifies a race car, and "a dipper" resembles an excavator.  Obviously, in the absence of additional information about the content of unknown words, the choice of "car" provides the least chance to make a mistake, because under the definition of "car" fall both a race car and an excavator. If we apply classical methods of classification, including the neural networks, to answer this question, all three options will be equally probable, as if the unknown part does not exist.

In academic literature, the problem of objects’ classification considering unknown features is not given enough attention. The paper of F. Provost and M. Saar-Tsechanski \cite{7} , in which the authors propose to remove unknown features from the training data and build an alternative model. In another paper \cite{5}, the authors propose to place the missing data in a specific feature called "absence". However, training the model in this way will resemble Laplace smoothing process, which is known to work well only in theory. Therefore, the existing methods do not provide a satisfactory result when working with natural text.

\begin{figure}[!h]
 \centering
\includegraphics[width=0.5\linewidth]{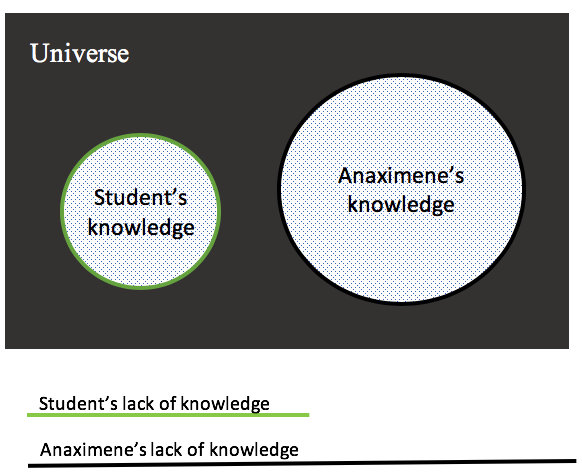}
\caption{Circles of Anaximene}
\label{image2}
\end{figure}

It would be worth mentioning two philosophical ideas that were considered in the development of the solution to the problem - the classification of the object only by its known features. The first of the ideas was presented by the Greek philosopher Anaximenes in the V century BC. Anaximenes while speculating about his knowledge and knowledge of his student, proposed the following formalism - the amount of knowledge was presented as the area of the circle, creating a border between the known and unknown knowledge (Fig. \ref{image2}). The smaller circle represented their knowledge of the student, and the larger – knowledge of Anaximene. Further, he draws attention to the fact that with increasing knowledge – the area of the circle, increases and lack of knowledge – the circumference. In the case of Anaximenes, it was larger, which meant his relative greater lack of knowledge.
\end{multicols}
\begin{multicols}{2}
The second idea is a "modern" logical continuation of the first. It was developed by the philosopher of science and founder of the third wave of positivism Karl Popper in his concept of verificationism \cite{2}. Popper argued that any hypothesis should allow the possibility of its verification and refutation by facts which could be still unknown at the time. An example of such facts is the existence of black swans, were discovered in 1697 on the Swan river in Australia, which refuted the old statement: "black swans do not occur in nature."

Taking this into account, we would like to introduce a new approach to the problem of classification with the use of neural networks, which allows to overcome the drawbacks in the case of classifying objects by known and unknown features. 

\section{Informational Neurobayesian Approach}

The basic method that we use to train neural networks is called Informational Neurobayesian Approach (INA) \cite{3} . Unlike traditional approaches to neural network training, which involve determining the weights of neural network features by a simple brute force method, which implies a high computational complexity of $O(n^2)$, INA allows to determine the weight coefficients of the neural network with significantly lower costs: by calculating (based on precedents) the initial approximation for typical descriptions of objects and selection for non-typical ones. This reduces the computational complexity of the enumeration problem to $O(nLog_n)$. Reducing complexity opens up opportunities for building large neuromodels (billions of parameters), previously available only for supercomputers and also to determine the required number of computational operations (resources) for the "training" of the neural network. Even more promising INA provides interpretability of weights as the amount of information. The first implication of this property is the ability to "understand" the precedents for which the neural network considers the chosen solution to be correct.  The second implication is the possibility of meaningful comparison of the output values of different neuromodels when selecting the resulting class.
\end{multicols}
\vspace{-0.5cm}
\begin{figure}[!h]
 \centering
\includegraphics[width=0.6\linewidth]{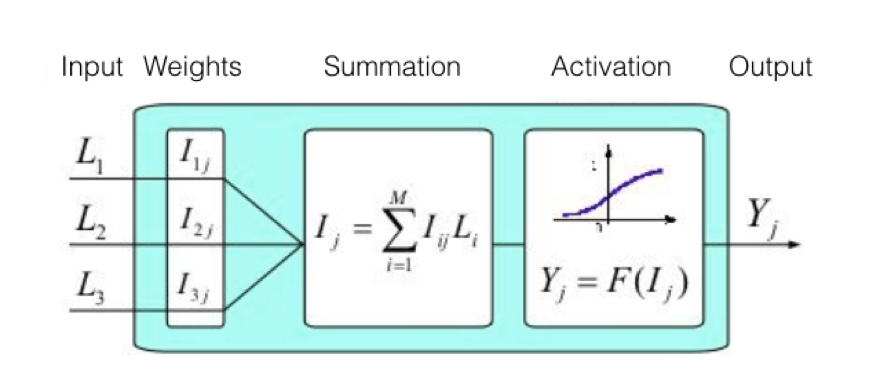}
\caption{A neuron in Informational Neurobayesian Approach}
\label{image3}
\end{figure}

\begin{multicols}{2}
The output of each neuron is the sum of the products of the weights of the connections and the input signal.

\begin{equation}
I_j= \sum w_{ij}x_i
\label{f1}
\end{equation}

The weight $w_{ij}\sim P_{ij}\ log_2\ P_{ij}$   the average quantity of pointwise mutual information (PMI) in the $i$-th feature for a $j$-class(object), determined on the basis of the increase in the probability of identification of the object of the $j$-th class, when belonging to the $i$-th feature. So $I_j=D_{KL} (P_{j\vert i} \vert\vert P_j)$.

Also, INA uses a special function of activation - AI, which determines the level of confidence (CU) of classification of the object to a given class. AI is the function with the highest value of 1, showing the exact correspondence of the object to a certain class and with the lowest value of -1 for the opposite case. AI is average for the following three functions
\begin{eqnarray*} 
AI=E \bigl[ SM(I_j)+ReLu(Norm\_of\_I_j)+  \\
  +Tanh(Weighted\_I_j)\bigr].
\end{eqnarray*}
Let us look more closely in AI components. \\
$SM(I_j)=\dfrac{e^{I_j}}{\sum_k e^{I_{k'}} }$ - SoftMax function, which represents the weight positive features of class $j$;

$ReLu(Norm\_of\_I_j)$ is a normalized function that represents the weight of these positive values for class j in relation to the sum of all positive values on all layers of the model.
\begin{multline*} 
\begin{matrix}
    ReLu(Norm\_I_j)& =
    & \left\{
    \begin{matrix}
    0 & \dfrac{I_j}{N_{features}} > l \\
    \dfrac{I_j}{N_{features}} & \mbox{else }
    \end{matrix} \right .
    \end{matrix} .
\end{multline*}
$Tanh(Weighted\_I_j)$ - the hyperbolic tangent function from the ratio of the features’ average weight describing an object, to the maximum average weight of one feature accurately describing a known object.

\begin{multline*} 
Tanh(Weighted\_I_j)=\\
= th \left(\dfrac{N_{features}\cdot I_j^{Max}}{MaxN_{features} \cdot I_j }\right).
\end{multline*}

\section{Problem and Solution}

In the paragraph below, we show how we have developed a solution to the problem of unknown features. 

\subsection{Problem Statement}

The proposed method was not compared with existing methods of neural network training as all of them have the same flaw – the more unknown features in the model – the worse is the forecast quality. It is important when we talk about language models as a few words can drastically change the meaning of a phrase. This problem has come from a practical task of using a robot to provide answers on government services. That is why this specific dataset has been used. 

Guided by the prerequisites set out in the introduction, we will look for a solution that guarantees the fulfillment of the following condition: the greater the proportion of unknown features relative to the known features, the less likely the correct classification of the object corresponding to the known features.

\begin{center}
\Large Known \& Unknown Features $\rightarrow$ \\ Object class
 
\normalsize  \textit{%Figure 3. 
Illustration of the connection of the object’s known and unknown features with its class}
\end{center}
If the weights of known features collectively exceed the weights of the unknown ones, the probability of correct determination of the class solely on the basis of the known features increases. Otherwise, it is practically impossible to determine the class only by known features.  Let us provide an example of this logic.

\begin{description}

\item[-]Unknown features have less weight than the known part.

\textit{"... has four wheels and a steering wheel unokuqhuba ngokukhawuleza \footnote{Here and after the language of Xhos is used for demonstration.} "- more likely that the correct class of the expression is "car".}

\item[-]Unknown features have more weight than the known part.

\textit{"... has four wheels and a steering wheel unako ukuphakamisa umthwalo omkhulu weebhakethi ngaphezu kweetoni eziliqela" - a lower probability that the correct class for the expression is "car".}

\end{description}

\subsection{Approach to the problem }

In accordance with the task in the INA framework it is necessary to find a mathematical generalization of the model \ref{f1}. Before proceeding to the mathematical formalization, we will form a Theorem that allows us to simplify the problem of this formalization.

\textbf{Theorem 1.} A model with limited knowledge. 

In conditions of limited data, an equal number of equally unknown features of an object cannot have different weights.

\textit{\textbf{Proof}}

As the amount of data is limited, the number of objects similar to a known object by certain features is also limited. In addition, the number of features describing these objects is limited as well. Therefore, the solution of the problem is reduced to the consideration of a finite number of objects that are as similar as possible on the given known features.  This means that if there is an object defined by the expression

\begin{equation*}
A=\underset{(\cdot)}{argmax} \sum_n w_{a_n,(\cdot)},
\end{equation*}

then for any similar object $\overset{\sim}{A}$ the following expression is true

\begin{equation}
\sum_n w_{a_n,\overset{\sim}{A}}=\sum_n w_{a_n,A},
\label{f2}
\end{equation}
where $a_n$ – is a feature description of the object A.

Let us consider the situation with one such object. Let there be a known object $A$ described by an array of features $ A =\{a_1,a_2,..a_n \}$. Also, there are two feature descriptions of objects $B$ and $C$. 

$$
B=\{a_1,a_2,..a_n,z_1^b,z_2^b..z_k^b \}
$$
$$
C=\{a_1,a_2,..a_n,z_1^c,z_2^c..z_k^c \}
$$

Containing part of $a_1,a_2,..a_n$ 
of object description and an unknown
$z_1^b,z_2^b..z_k^b$ and $z_1^c,z_2^c..z_k^c$ respectively.

Let us assume that the sum of feature weights of objects $B$ and $C$ for class (object) $A$ is not equal, i.e.

\begin{equation}
\sum_C w_{(\cdot),A} \neq \sum_B w_{k,A}.
\label{f3}
\end{equation}

Such a condition would mean that one of the objects from $B$ and $C$ is more similar to $A$ 

\begin{equation*}
\sum_C w_k,A =\sum_n w_{a_n,A}+\sum_k w_{z_k^c,A}
\end{equation*}
\begin{equation*}
\sum_B w_k,A =\sum_n w_{a_n,A}+\sum_k w_{z_k^b,A}
\end{equation*}
then expression (\ref{f3}) is equivalent to 

\begin{equation}
\sum_k w_{z_k^c,A} \neq \sum_k w_{z_k^b,A}.
\label{f4}
\end{equation}

But since the unknown parts have the same number k of unknown features, they are not equally known, that is

\begin{equation*}
\sum_k w_k^c,A  \sim \sum_n w_{a_n,A}
\end{equation*}
and 
\begin{equation*}
\sum_k w_k^b,A  \sim k\sum_n w_{a_n,A}
\end{equation*}

And since each unknown feature of $B$ and $C$ is equally unknown with respect to each feature of the object under consideration – $A$, then

\begin{multline}
\lbrace  \forall A, B, C\ \ \exists \alpha_A : \sum_k w_{z_k^c,A}=\\
=\sum_k w_{z_k^b,A}=\alpha_A k\sum_n w_{a_n,A} \rbrace 
\label{f5}
\end{multline}

where $\alpha_A$ is a fixed coefficient of proportionality for the sum of weights of features of an object $A$. 

The expression (\ref{f5}) leads us to a contradiction in statement (\ref{f4}), and accordingly, makes the statement (\ref{f3}) incorrect. Because for any arbitrary known object in the group that is as similar as possible, given (0) we obtain

\begin{equation*}
k\sum_n w_{a_n,\overset{\sim}{A}}=k\sum_n w_{a_n,A}
\end{equation*}

So that expression (\ref{f4}) remains true with limited data. As was to be proved.

From Theorem 1 we can draw the following conclusion:

\textbf{Conclusion:} In conditions of limited data, fewer equally unknown features of an object cannot have different weights.

In accordance with theoretical conclusions within the framework of the task, a necessary and sufficient solution is a method that provides a change in the output neuron signal in proportion to the number of unknown features and the total weight of the known features. The mathematical formalization of this solution is the modified expression (\ref{f3})
\begin{equation}
I_j=\left( 1-\alpha_f \dfrac{N_{unknown features}}{1+N_{unknown features}}\right)\sum w_{ij}x_i
\label{f6}
\end{equation}

The presented model of discounting weights (\ref{f6}) is universal and, on account of $\alpha_j$ coefficient, can be adapted for different activation functions.

\section{Practical application}

The presented model and the theorem have been successfully applied to create a question-answering system. The task was to develop a question-and-answer system based on AI, which would select the type of service based on freely formulated user requests.

The original dataset contained records of more than 1,500 user queries collected during the tests. The requests were divided into approximately 400 different categories. 

The process of development was conducted in seven stages.

\begin{description}
\item[1]Elimination of punctuation.
\item[2]Correction of spelling errors.
\item[3]Elimination of duplicate words.
\item[4]Making a list of synonyms (the list of synonyms was compiled manually for the main words and expressions).
\item[5] Queries have been converted according to the list of synonyms. 
The provided data set was too small to provide answers with a high level of confidence – the model contained a small number of words. A classic approach to solving this problem would be to compile a Word2vec \cite{6}  model trained on a large data set. However, the real queries of people contain a large number of colloquial, informal vocabulary and abbreviations, which makes the compilation of the relevant data set too resource and time-consuming. The compilation of the synonym dictionary took only about three days, while it would take at least a month to compile a data set with similar or less accuracy at the output.

\item[6]Words have been lemmatized through the application of our own design - a Big Linguistic Model, producing the grammatical and the lexical markup of the texts, transforming words into unigrams – initial word forms. This allows with only a small dataset to train the model to recognize a much larger number of words.

\item[7] In addition to unigrams, we have also selected their meaningful pairs from the queries. For example, "what's your name": "what+your", "what+name", "name+ your".
\end{description}

Tab \ref{tab1} shows an example from the dataset after it has been preprocessed. 

\end{multicols}

\begin{table}[h]
\caption{A row from the dataset after preprocessing}
\begin{tabular}{lllll}
\hline
\multicolumn{1}{|l|}{}         & \multicolumn{1}{c|}{Query}                                                                                                                                                                                       & \multicolumn{1}{l|}{Type of question}      & \multicolumn{1}{l|}{Lel of question}            & \multicolumn{1}{l|}{Category}                                \\ \hline
\multicolumn{1}{|l|}{Unigrams} & \multicolumn{1}{l|}{can I have several international passports}                                                                                                                                                  & \multicolumn{1}{c|}{\multirow{2}{*}{Info}} & \multicolumn{1}{c|}{\multirow{2}{*}{Service 6.0}} & \multicolumn{1}{c|}{\multirow{2}{*}{In\_passport\_info\_21}} \\ \cline{1-2}
\multicolumn{1}{|l|}{Bigrams}  & \multicolumn{1}{l|}{\begin{tabular}[c]{@{}l@{}}can+I'\\ 'I+have'\\ 'have+several'\\ 'international+passports'\\ 'several+passports'\\ 'have+passports'\\ 'several+intternational'\\ 'have+several'\end{tabular}} & \multicolumn{1}{c|}{}                      & \multicolumn{1}{c|}{}                             & \multicolumn{1}{c|}{}                                        \\ \hline                                                           
\label{tab1} 
\end{tabular}
\end{table}
\begin{multicols}{2}
After training, the model is ready to automatically process queries in the same 7 steps, which include preparing the dataset as well as two additional steps:
\begin{description}
\item[-]Classification of the processed query by the model that matched query bigrams and unigrams with those present in the model.
\item[-] Assessment of the level of confidence (CL) of the model in the correctness of classification from -1 to 1.
\end{description}
The bigrams and unigrams included in the model, as well as in the queries, are known features with a positive weight that increase the CL, while the rest of the words present in the query, but are not recognized by the model are unknown values with a negative weight that lower the CL of the model.

Trained with this method, the model showed results of 0.8 precision, which is a good result, but insufficient for the task. Accordingly, in order to improve its functioning, we decided to abandon the spell heck of user queries due to unjustified transformation of unknown words to known ones, which generally gives a worse result. In addition, it was necessary to solve several issues:

\begin{description}
\item[A]Solve the problem of correspondence of modified queries to several classes with the same level of confidence. After 16 iterations, an acceptable level of confidence was achieved, which implies a correct answer with CL above 0.6, and incorrect - below 0.6. 
\item[B]Make sure that most of the modified queries are correctly classified with a confidence level greater than 0.5 (50\%).

As a result, a solution was presented in which the user is shown up to 5 closest original queries from different classes in the confidence range of 5\% and is asked to choose the most appropriate one. 
\item[C]Make sure that most of the non-relevant requests are classified with a confidence level below 0.5 (50\%). This problem was of the greatest importance. Irrelevant queries in most cases contained a number of common words recognized by the model. In the case of typical Softmax functions, such requests were classified with a high level of confidence, which is unacceptable in this case.
\end{description}
Practical testing has led us to the conclusion that considering the dictionary of synonyms, queries that contain more than a third of unknown words, with a high probability are irrelevant. This allowed us to determine the minimum confidence threshold of 0.6. 
\end{multicols}
\begin{table}[!ht]
\centering
\caption{Model quality}
\begin{tabular}{lllrl}
\cline{1-4}
\multicolumn{1}{|c|}{\textbf{Model}} & \multicolumn{1}{c|}{\textbf{Test Data}} & \multicolumn{1}{c|}{\textbf{Method}} & \multicolumn{1}{c|}{\textbf{Accuracy}} &  \\ \cline{1-4}
\multicolumn{1}{|l|}{\multirow{4}{*}{\begin{tabular}[c]{@{}l@{}}Bigrams and unigrams\\  are features, sentence\\  ID is class\end{tabular}}} & \multicolumn{1}{l|}{\multirow{2}{*}{\begin{tabular}[c]{@{}l@{}}The similar as in a train \\ dataset (only known words)\end{tabular}}} & \multicolumn{1}{l|}{Basic INA} & \multicolumn{1}{r|}{0.9364} & \multicolumn{1}{c}{} \\ \cline{3-4}
\multicolumn{1}{|l|}{} & \multicolumn{1}{l|}{} & \multicolumn{1}{l|}{Updated INA} & \multicolumn{1}{r|}{0.9364} &  \\ \cline{2-4}
\multicolumn{1}{|l|}{} & \multicolumn{1}{l|}{\multirow{2}{*}{\begin{tabular}[c]{@{}l@{}}Half rows consist of\\  unknown words\end{tabular}}} & \multicolumn{1}{l|}{Basic INA} & \multicolumn{1}{r|}{0.4330} &  \\ \cline{3-4}
\multicolumn{1}{|l|}{} & \multicolumn{1}{l|}{} & \multicolumn{1}{l|}{Updated INA} & \multicolumn{1}{r|}{0.9364} &  \\ \cline{1-4}
 &  &  & \multicolumn{1}{l}{} & 
\label{tab2}
\end{tabular}
\end{table}

\begin{figure}[!h]
 \centering
\includegraphics[width=0.7\linewidth]{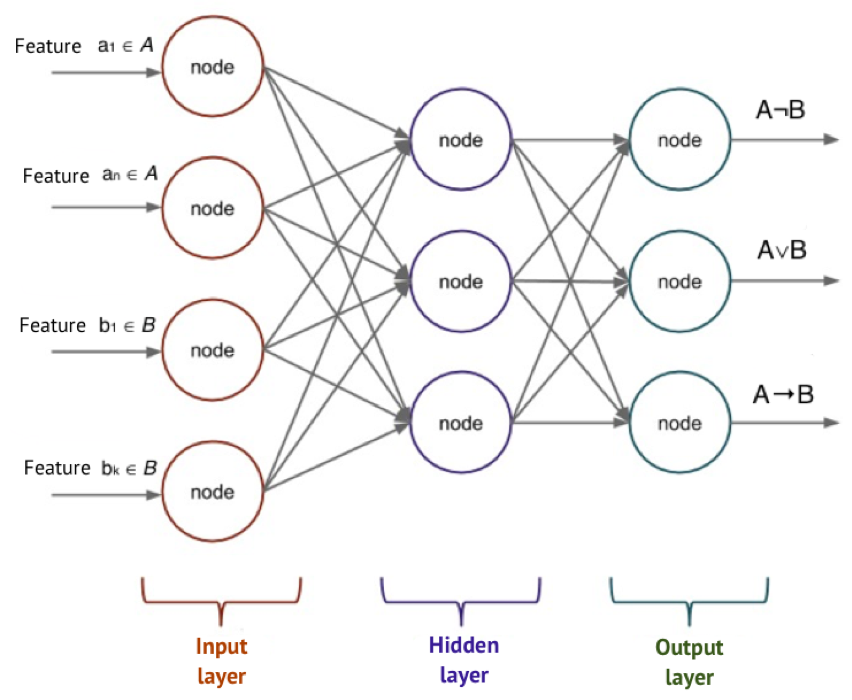}
\caption{Illustration of approach to classification of unknown objects by selection of comparison method of unknown object’s features with features of known objects}
\label{image6}
\end{figure}

\begin{multicols}{2}
\section{Further development}

Despite the high accuracy results of the model trained with the help of the described approach, in practice there is a need for the classification of objects not only by partially known features, but also classification of previously unknown by the model objects. The statement of the problem seems to be contradictory, since a typical classification using a neuromodel involves a choice among the classes of objects on which the model was trained on. For example, in the task with classification of vehicles, we can imagine that in the presence of objects (excavator, race car, and car) with known features in the model appears a new object – a bicycle with known features – a steering wheel and wheels, and unknown features – pedals and chain. In this case, the bicycle could be attributed by the model to one of the three known objects, which would be a mistake. It is also clear that we are interested in a solution that does not require changing the model classes, that is, retraining. And yet this problem has a solution that seems to us so relevant and new that we decided to present it at the end of this article in the most general form.

We propose to classify not the object itself, but the method of the model’s decision-making - how to process the input data. In this case, the final goals of this processing can be different, including the solution to the problem of query classification by selecting the method of evaluation of semantic proximity to the queries of known classes. It is important that the methods of processing, apparently, are reduced to the basic operations of the logic algebra – negation, conjunction, disjunction, implication, dual stroke, etc.

Thus, the known features of unknown objects become the initial data on the basis of which the choice (using a neuromodel) of the method of data processing is made. It is also important that this process retains the possibility of operational (in the process of using) "additional training" of the neuromodel by gradient descent or simulated annealing of the neural network weights, which in case of a choice that does not allow to obtain the correct answer of the data processing method.

In the future, the authors of the article plan to test this approach and, if successful, to supplement the proposed approach of objects’ classification considering unknown features with the possibility of solving the problem of unknown classes.

\section*{Conclusion}
The results of this work can be summarized as follows.
\begin{description}
\item[1] A new algorithm of neural network, which considers not only the sum of known features’ weights, but also the sum of unknown ones.
\item[2] The theorem "On the significance of unknown features in the classification of objects" was formulated and proved.
\item[3] A model of a question-answer system for the analysis and classification of user queries was trained. For this the authors used their own Informational Neurobayesian Approach, according to which the weights of the model are presented in the form of interpretable bits of information. 
\item[4] Preparation of a synonym set for the keywords present in the dataset can significantly reduce the time spent on model development ensuring the classification quality of the queries above 0.9, including when using alternative formulations of queries.
\item[5]The use of not only single words (unigrams) as features, but also related pairs of words (bigrams) allows to increase the relevance of the model answers.
\item[6] The algorithm for determining the weight of unknown features proved to be effective for filtering irrelevant queries, increased the accuracy of forecasts up to 0.936 without reducing recall.
\item[7] The presented method proved to be highly effective in the case of development of question-answer systems trained on a limited number of phrases and are further used to answer freely formulated questions containing words and phrases that were not present in the original data set.
\item[8] The future development of the approach will provide the ability of the model to classify objects not only on the part of the well-known features, but even the to classify unknown (model) classes of objects.
\end{description}

\bibliographystyle{unsrt}

\end{multicols}

\end{document}